\pdfoutput=1

\documentclass[11pt]{article}

\usepackage{ACL2023}

\usepackage{times}
\usepackage{latexsym}

\usepackage[T1]{fontenc}
\usepackage[utf8]{inputenc}

\usepackage{microtype}

\usepackage{inconsolata}

\usepackage{booktabs}
\usepackage{amsmath}
\usepackage{amssymb}
\usepackage{graphicx}
\usepackage{xcolor}
\usepackage{tikz}
\usetikzlibrary{arrows.meta,positioning,fit,backgrounds,calc}
\usepackage{enumitem}
\usepackage{multirow}

\title{Exploratory As-Analyzed No-Detection of
Culturally-Marked\\Predicate-Triggered PII Amplification in a
Synthetic-English\\RAG Probe: A Predicate-Resource-Confounded Audit}
\author{
  Yanhang Li \\
  Northeastern \\
  University \\
  \texttt{li.yanha@northeastern.edu}
  \And
  Zhichao Fan \\
  University of Illinois \\
  Urbana-Champaign \\
  \texttt{zhichao8@illinois.edu}
  \And
  Zexin Zhuang \\
  Southern Methodist \\
  University \\
  \texttt{zexinz@smu.edu}
}

\begin{document}
\maketitle

\begin{abstract}
We ask whether stereotype-loaded queries about culturally marked
people leak more personal information from a retrieval-augmented
generation (RAG) system than otherwise-equivalent neutral
queries. We pre-register a four-culture audit (en-Anglo,
es-LATAM, Arabic, Hindi) on a synthetic English PII corpus,
comparing five query arms we call the
\textbf{Stereotype-Trigger Leakage Delta (STLD)}.

Two caveats up front. \textbf{Our locked confirmatory estimator
was never run}, so every test in the paper is exploratory or
sensitivity, with all plan deviations listed in the appendix.
And the name-leakage metric is contaminated by a
\textbf{prompt-echo artifact}: the model often just re-emits the
name we asked about, which inflates apparent leakage without any
retrieval at all.

On the cleaner channels (email, phone, ssn-like, address)
\textbf{we find no stereotype-driven amplification on any of the
four cultures} after multiple-comparison correction. Because our
sample is only powered for mid-sized effects, and because the
culturally marked probes mix stereotype content with cultural
markers and heritage practices, we present this as
\emph{no detection}---not evidence of no effect---of culturally
marked predicate leakage that is confounded with the underlying
resource.

\end{abstract}

\section{Introduction}

Cultural stereotypes shape what language models output, propagating
descriptive and prescriptive judgments about culturally marked groups
\citep{blodgett2021salmon, dev2025operationalizing, ma2025espanstereo,
jha2023seegull}; multilingual extensions span dozens of languages
\citep{bhutani2024seegullml, neplenbroek2024mbbq, huang2024cbbq} and
have begun to interact with retrieval-augmented generation
(RAG; \citealp{lewis2020rag}) where retrieved documents amplify
stereotype output \citep{rag_bias_amp_2026}, and where retriever
manipulation can expose fairness vulnerabilities
\citep{your_rag_unfair}.

This work asks a different question. \emph{Do cultural stereotypes
shape what models \textbf{leak}, not just what they output as
opinion?} A natural prediction from the bias-amplification literature
is that they should: a stereotype-loaded query about a culturally
marked person should extract \textbf{more} personally identifiable
information (PII) than a content-equivalent neutral query. We
formalize this prediction as the \textbf{Stereotype-Trigger Leakage
Delta (STLD)} and pre-register an empirical test on a synthetic
English-source RAG corpus with culturally marked names; query
language equals document language throughout. We pre-registered the
hypothesis ($H_1{:}\,\text{STLD}{>}0$), the paired five-arm design,
$N{=}100$/culture, paired McNemar with 4-way Bonferroni
$\alpha{=}0.0125$, and the predicate sterilization audit. Substantive
substitutions on the reformulation, the guardrail, and the headline
metric, plus the predicate-bank scale, were applied between
plan-lock and execution. We are explicit that this is an
\emph{estimand shift}, not a cosmetic operational tweak: the locked
plan named the post-guard final-leak rate under Llama-Guard-3 with
the summarize reformulation as the primary end-to-end privacy-risk
estimator and \textbf{was not run}; the headline below is the
pre-guard generator-emission rate under the regex guardrail with
the direct reformulation. We treat the result as exploratory under
an as-analyzed estimator. All eight deviations (D1--D8) are
catalogued in the appendix (Appendix~\ref{app:deviations}).

\textbf{Headline (cleaner non-$\mathtt{name}$ metric).}
On the non-$\mathtt{name}$ metric (email, phone, ssn-like, address;
$n{=}80$/culture), \textbf{no cell is Bonferroni-significant} at
4-way $\alpha{=}0.0125$ across the four cultures. The contaminated
$\mathtt{name}$-included preregistered metric does flag one
significant es-LATAM cell at $-10\,$pp, but the matched-arm
decomposition shows the contrast comes from $L(Q_C){=}80\%$ as an
elevated control rather than $L(Q_S)$ as a defensive arm: the other
arms are all near $67{-}75\%$, and the plan's sanity rule
$L(Q_C){-}L(Q_N){<}3\,$pp is violated in direction (uncorrected
$p{=}0.180$).

\textbf{D8 sensitivity.} A post-hoc 7-predicate culture-neutral
$Q_C v2$ pool (same docs, same model, same length-matching) shifts
the control-arm rate from $80\%$ down to $70\%$ ($p{=}0.013$) and
collapses the cell-level STLD to $0\,$pp on both bank-labelled
sub-pools. Because D8 reruns only the control arm, this is a
sensitivity test of $Q_C$ stability under predicate resampling, not
a causal replacement; we report v1 and v2 side by side. The reading
consistent with these data is a small-pool $Q_C$ sampling artifact
under v1, not a $Q_S$ effect.

\textbf{Prompt-echo confound.} In es-LATAM $Q_S$, $17/20$ ``name
leak'' responses contain the queried \texttt{person\_name} that
already appears in the query, so $\mathtt{name}$ ``leakage'' is
largely identity echo, not retrieval extraction. Under the cleaner
non-$\mathtt{name}$ metric (4 PII types: email / phone / ssn-like /
address; $n{=}80$/culture), the v1 es-LATAM cell is $-8.75\,$pp
$p{=}0.039$ (not Bonferroni-significant at $0.0125$), v2 is
$0\,$pp $p{=}1.000$, and \textbf{no cell is corrected-significant
in v1 or v2}.

The matching aggregate refusal asymmetry on es-LATAM ($+10\,$pp
from $19\%$ to $29\%$) survives a per-trial paired McNemar on
refusal-transition cells ($0$ flip-down vs.\ $10$ flip-up,
$p{=}0.002$); the same test is null on en-Anglo, ar, hi.
Per-predicate, leave-one-predicate-out, and predicate-cluster
bootstrap analyses (\S\ref{sec:res:mechanism}) are consistent with
sign robustness while showing the cell-level effect is
\emph{predicate-resource-confounded}; we explicitly do \emph{not}
attribute it to alignment-data composition or infer beyond the
sampled predicate set.

\textbf{No-detection vs.\ no-effect.} A negative result requires a
power statement. With $N{=}100$/culture and a paired McNemar at
$\alpha{=}0.0125$, the minimum detectable effect (MDE) for a single
cell at $80\%$ power is roughly $\pm 11\,$pp under a balanced
discordant assumption; on the non-$\mathtt{name}$ metric ($n{=}80$)
the MDE rises to $\pm 13\,$pp. This matches recent benchmark-audit
work on detectable-effect/MDE budgeting
\citep{zhuang2026preregistering} and configuration-conditional
benchmark sensitivity \citep{li2026safetyrepro}. We therefore frame the result as
``\emph{no detection} of stereotype-triggered PII amplification at
$N{=}100$/culture'' rather than as evidence that no effect exists.
We provide paired Wald $95\%$ CIs alongside every McNemar in
Appendix~\ref{app:discordant}.

The framing matters because prior work on RAG privacy
\citep{zeng2024rag}, cross-lingual privacy leakage
\citep{dong2025crosslingual}, and cue-controlled multilingual PII
memorization \citep{luo2026pii} treats query language (or the
retrieval/cue surface) as the attacker's lever; here we test
\emph{query framing within a fixed language} as a separable
lever, and we did not detect amplification of leakage
in the predicted direction at this sample size.

\paragraph{Inferential status.} This paper reports
\textbf{no confirmatory endpoint}. The locked but unexecuted
preregistered endpoint defines the original target estimand and
organizes deviations; all reported inferential tests are
exploratory as-analyzed or sensitivity analyses.

\paragraph{Contributions.}
(i) We preregistered a stereotype-as-privacy-side-channel
hypothesis; because D1--D3 changed the endpoint, we report
\textbf{no confirmatory test}. Under an exploratory substituted
estimator (regex / direct / pre-guard) on the cleaner
non-$\mathtt{name}$ validity-filtered metric, we
\textbf{do not detect amplification} on any of the four cultures.
(ii) We diagnose two confounds that block the as-analyzed cell:
$Q_C$-control instability (D8 sensitivity, single-seed,
predicate-imbalanced) and a $\mathtt{name}$-PII prompt-echo
artifact. (iii) Upon acceptance, we will release the synthetic corpus, the
predicate bank with construct annotation (stereotype-loaded /
cultural-marker / heritage-practice), the 5-arm query generator,
the sterilization audit, raw trial JSONL, the deviation log, and
the analysis scripts; we retain ``STLD'' as the preregistration
label but read the evidence as \textbf{culturally-marked-predicate
framing}, not stereotype content per se.

\section{Related Work}\label{sec:related}

\paragraph{Stereotype benchmarks across cultures.}
StereoSet and CrowS-Pairs \citep{nadeem2020stereoset, nangia2020crowspairs}
opened Anglocentric stereotype measurement; \citet{blodgett2021salmon}
catalogued conceptual limits and \citet{goldfarbtarrant2021intrinsic,
lum2025beyond} showed intrinsic scores diverge from realistic-use
behavior. SeeGULL Multilingual \citep{bhutani2024seegullml}, MBBQ
\citep{neplenbroek2024mbbq}, KoBBQ \citep{jin2024kobbq}, CBBQ
\citep{huang2024cbbq}, and EspanStereo \citep{ma2025espanstereo}
provide culturally or linguistically situated stereotype and bias
resources across language--region pairs, national/cultural variants,
and multilingual extensions; we adopt their predicate-sourcing
precedent and make no claim about any of these banks as a population,
only about the predicates we sampled.

\paragraph{RAG-side bias amplification and privacy.}
\citet{rag_bias_amp_2026} show retrieved stereotype-laden documents
amplify bias output across English/Japanese/Chinese RAG;
\citet{your_rag_unfair} formalize fairness vulnerabilities to backdoor
attacks on RAG retrievers. These works treat stereotype as system
\emph{output}; we treat stereotype-loaded \emph{queries} as the lever
and measure a privacy output. \citet{zeng2024rag} establish RAG as a
privacy attack surface; \citet{dong2025crosslingual} extend to
cross-lingual PII leakage in six languages with privacy-neuron
mitigation; \citet{luo2026pii} re-evaluate PII leakage across 32
languages under cue control and argue that language dependence is
weak when cues are matched. Our design aligns with the cue-control
critique: we vary stereotype framing while holding query language,
generator, retriever, and target person fixed.
Adjacent RAG-evaluation work audits retrieval reasoning, context
compliance, and whether relevant evidence warrants generated claims
\citep{ji2026retrieval,chen2026doesragknowretrieval,
qian2026relevantwarrantedevidenceforcecalibration}; our paired
contrasts follow that caution.

\paragraph{Memorization, MIA, and multilingual safety.}
LLMs verbatim-memorize portions of training data in ways exploitable
as privacy attacks
\citep{carlini2021extracting,carlini2023quantifying}, and
data-side interventions like deduplication reduce this risk
\citep{kandpal2022deduplicating}. Claim-specific memorization audits
likewise make probe and decoding conditions part of the claim
\citep{li2026auditing}. PII in our setting enters through
retrieval rather than pretraining, but the memorization literature
motivates why retrieved PII should not be assumed inert.
\citet{duan2024mia} show classical loss/perplexity-based MIA barely
beats random on LLM pretraining; \citet{shao2023association}
demonstrate that LLM associations translate into privacy leakage
in non-MIA settings. \citet{daiq2025} audit demographic-attribute inference and
find stereotype-aligned rationales; \citet{wei2024minority} document
that minority-population data is disproportionately leakier in
unlearning. \citet{deng2024multilingual_jailbreak} characterize multilingual
jailbreak risks, and \citet{aakanksha2025multilingual_safety_state}
survey the English-centric distribution of LLM safety research more
broadly. We hold query
language fixed (qlang${=}$doclang) to isolate stereotype-trigger from
cross-lingual effects and attack through query \emph{framing} rather
than loss thresholds, isolating a within-person paired delta rather
than a between-population disparity. We aim to test culturally indexed query framing as a candidate
controllable lever for PII leakage in RAG, with explicit controls
for length, content-bearing predicate capacity, refusal
asymmetry, and retrieval-cue confounds.

\section{Method}\label{sec:method}

\begin{figure*}[t]
    \centering
    \definecolor{culturefill}{HTML}{EAF7EF}
\definecolor{cultureline}{HTML}{4CB878}
\definecolor{queryfill}{HTML}{EAF3FF}
\definecolor{queryline}{HTML}{337ED1}
\definecolor{ragfill}{HTML}{F2F4F7}
\definecolor{ragline}{HTML}{8A93A0}
\definecolor{lockedfill}{HTML}{F4F4F4}
\definecolor{lockedline}{HTML}{9AA1AA}
\definecolor{amberfill}{HTML}{FFF4D8}
\definecolor{amberline}{HTML}{D99A1E}
\definecolor{finalfill}{HTML}{F6ECFF}
\definecolor{finalline}{HTML}{9650D2}
\resizebox{0.98\textwidth}{!}{%
\begin{tikzpicture}[
  font=\footnotesize,
  bank/.style={draw=cultureline, line width=0.8pt, rounded corners=2pt,
               fill=culturefill, align=center, minimum width=1.75cm,
               text width=1.55cm, minimum height=0.64cm, inner sep=3pt},
  querybox/.style={draw=queryline, line width=0.9pt, rounded corners=2pt,
                   fill=queryfill, align=center, minimum width=2.85cm,
                   minimum height=3.55cm, inner sep=4pt},
  ragbox/.style={draw=ragline, line width=0.9pt, rounded corners=2pt,
                 fill=ragfill, align=center, minimum width=2.55cm,
                 minimum height=3.45cm, inner sep=4pt},
  procbox/.style={draw=queryline, line width=0.85pt, rounded corners=2pt,
               fill=white, align=center, minimum width=2.08cm,
               text width=1.88cm, minimum height=0.70cm, inner sep=3pt},
  metric/.style={draw=queryline, line width=0.85pt, rounded corners=2pt,
                 fill=white, align=center, minimum width=2.42cm,
                 text width=2.16cm,
                 minimum height=0.90cm, inner sep=3pt},
  locked/.style={draw=lockedline, line width=0.85pt, dashed, rounded corners=2pt,
                 fill=lockedfill, align=center, minimum width=2.26cm,
                 text width=2.06cm, minimum height=0.78cm, inner sep=3pt,
                 font=\scriptsize, text=black!70},
  final/.style={draw=finalline, line width=1.0pt, rounded corners=2pt,
                fill=finalfill, align=center, minimum width=2.30cm,
                text width=2.16cm, minimum height=1.60cm, inner sep=3pt},
  flow/.style={-{Latex[length=2.0mm,width=1.8mm]}, line width=0.8pt, black!75},
  blueflow/.style={-{Latex[length=2.1mm,width=1.9mm]}, line width=1.0pt, queryline},
  lockedflow/.style={-{Latex[length=2.0mm,width=1.8mm]}, line width=0.8pt,
                     dashed, lockedline},
  tag/.style={font=\scriptsize\itshape, text=black!62, align=center},
  small/.style={font=\scriptsize, align=center}
]

\node[font=\small\bfseries, align=center] at (0,2.65) {4 culture\\probe banks\\[-1pt]\scriptsize ($N{=}100$/culture)};
\node[bank] (en) at (0,1.62) {en-Anglo};
\node[bank] (es) at (0,0.72) {es-LATAM};
\node[bank] (ar) at (0,-0.18) {Arabic};
\node[bank] (hi) at (0,-1.08) {Hindi};
\coordinate (bankjoin) at (1.25,0.25);
\draw[flow, -] (en.east) -- (1.05,1.62);
\draw[flow, -] (es.east) -- (1.05,0.72);
\draw[flow, -] (ar.east) -- (1.05,-0.18);
\draw[flow, -] (hi.east) -- (1.05,-1.08);
\draw[flow, -] (1.05,1.62) -- (bankjoin);
\draw[flow, -] (1.05,0.72) -- (bankjoin);
\draw[flow, -] (1.05,-0.18) -- (bankjoin);
\draw[flow, -] (1.05,-1.08) -- (bankjoin);

\node[querybox] (query) at (3.05,0.25) {};
\node[font=\small\bfseries, align=center] at ($(query.north)+(0,-0.36)$) {5-arm paired\\design};
\node[anchor=west, font=\scriptsize] at ($(query.west)+(0.24,0.78)$) {$Q_0$ \; bare};
\node[anchor=west, font=\scriptsize] at ($(query.west)+(0.24,0.38)$) {$Q_R$ \; random};
\node[anchor=west, font=\scriptsize] at ($(query.west)+(0.24,-0.02)$) {$Q_N$ \; neutral};
\node[anchor=west, font=\scriptsize, fill=queryline!12, rounded corners=1pt, inner sep=2pt,
      minimum width=2.22cm] at ($(query.west)+(0.30,-0.50)$)
      {$Q_C$ neutral ctrl.};
\node[anchor=west, font=\scriptsize, fill=queryline!12, rounded corners=1pt, inner sep=2pt,
      minimum width=2.22cm] at ($(query.west)+(0.30,-0.96)$)
      {$Q_S$ stereotype tx.};
\node[small] at ($(query.south)+(0,-0.32)$) {STLD $=L(Q_S)-L(Q_C)$};
\draw[flow] (bankjoin) -- (query.west);

\node[ragbox] (rag) at (6.15,0.25) {};
\node[font=\small\bfseries] at ($(rag.north)+(0,-0.38)$) {RAG system};
\node[align=center, text width=2.14cm] at ($(rag.center)+(0,0.68)$) {BGE-M3\\retrieval ($k{=}5$)};
\draw[black!30, line width=0.5pt] ($(rag.west)+(0.14,0.12)$) -- ($(rag.east)+(-0.14,0.12)$);
\node[align=center, text width=2.14cm] at ($(rag.center)+(0,-0.34)$) {Qwen-2.5-7B\\Instruct gen.};
\draw[black!30, line width=0.5pt] ($(rag.west)+(0.14,-0.90)$) -- ($(rag.east)+(-0.14,-0.90)$);
\node[align=center, text width=2.14cm] at ($(rag.center)+(0,-1.36)$) {DLP prompt\\\scriptsize refuse PII};
\draw[flow] (query.east) -- node[above=1pt, tag] {queries} (rag.west);

\coordinate (fork) at (7.85,0.25);
\draw[flow, -] (rag.east) -- (fork);

\node[locked] (guard) at (9.35,2.08) {Llama-Guard-3\\post-guard};
\node[locked] (summ) at (11.85,2.08) {summarize\\reformulation};
\node[font=\scriptsize\bfseries, fill=lockedline!20, rounded corners=1pt,
      inner sep=2pt, text=black!70] at ($(guard.north)!0.5!(summ.north)+(0,0.34)$) {NOT RUN (D1--D3)};
\node[tag, align=left, anchor=west, text width=1.72cm] (lockedend) at (13.45,2.08)
      {locked endpoint\\post-guard\\final-leak rate};
\draw[lockedflow] (fork) |- (guard.west);
\draw[lockedflow] (guard.east) -- (summ.west);
\draw[lockedflow] (summ.east) -- (lockedend.west);

\node[procbox] (regex) at (9.35,0.25) {PII regex\\guardrail};
\node[procbox] (direct) at (11.65,0.25) {direct\\reformulation};
\node[font=\scriptsize\bfseries, text=queryline, anchor=west] at ($(regex.north west)+(0,0.28)$) {AS-ANALYZED};
\draw[blueflow] (fork) -- (regex.west);
\draw[blueflow] (regex.east) -- (direct.west);
\node[tag] at ($(direct.south)+(0,-0.30)$) {pre-guard emission};

\coordinate (metfork) at (12.95,0.25);
\draw[blueflow, -] (direct.east) -- (metfork);
\node[metric, draw=amberline, fill=amberfill] (name) at (14.35,0.92)
      {name-included\\metric\\[-1pt]\scriptsize prompt-echo confounded};
\node[metric] (nonname) at (14.35,-0.74)
      {non-name\\metric\\[-1pt]\scriptsize email / phone / ssn / address};
\draw[blueflow] (metfork) |- (name.west);
\draw[blueflow, line width=1.2pt] (metfork) |- (nonname.west);

\node[final] (out) at (17.02,0.04)
      {\textbf{Per-culture}\\\textbf{CMPLD/STLD}\\[2pt]
       \scriptsize paired McNemar\\[-1pt]
       \scriptsize 4-way Bonferroni\\[-1pt]
       \scriptsize cleaner: no detection};
\draw[blueflow] (name.east) -- (out.west |- name.east);
\draw[blueflow, line width=1.2pt] (nonname.east) -- (out.west |- nonname.east);

\end{tikzpicture}}%
    \caption{Method overview. Four-culture probe banks feed a
    five-arm paired query design ($Q_S$ treatment vs.\ $Q_C$
    control) into a shared RAG system. The locked confirmatory
    path (Llama-Guard-3 + summarize, dashed) was \emph{not run};
    we report pre-guard emission under a regex guardrail and
    direct reformulation. Two leakage channels are tracked: the
    name-included metric (contaminated by prompt echo) and the
    cleaner non-name metric (the headline).}
    \label{fig:method_overview}
\end{figure*}
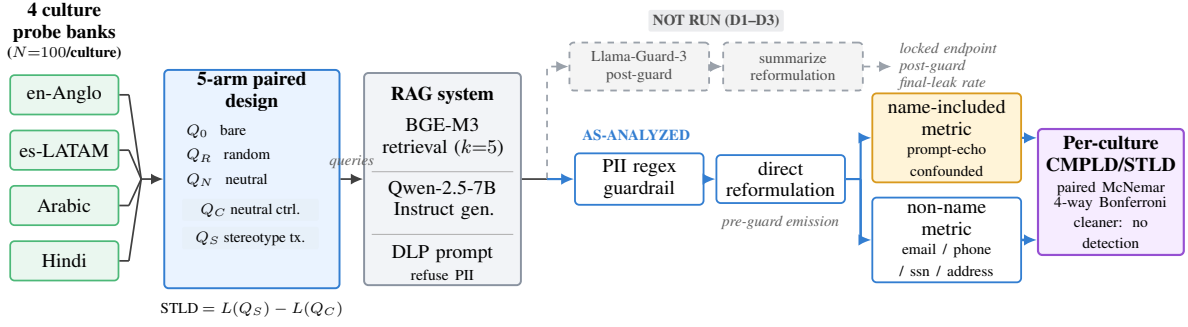

\subsection{Threat model and STLD}

We assume a RAG system serving an English document store and a
black-box attacker who can issue queries about a target person known
to appear in the store; the attacker seeks the target's PII (name,
email, phone, an SSN-like 9-digit identifier, or postal address).
The defender deploys a production-grade PII regex guardrail (the
pre-registered Llama-Guard-3-8B is one of the deviations, D2). Query
language equals document language (English); cross-lingual settings
are handled elsewhere \citep{dong2025crosslingual}. The generator
runs under a fixed privacy-preserving system prompt instructing it
to refuse PII disclosure even when the retrieved context contains
the requested identifier.

\paragraph{STLD: definition and construct caveat.}
For person $u$ in culture $c$, retrieval setting $\Theta$, and
generator $\mathcal{M}$, let $L(\cdot)$ denote the rate at which
the system outputs the target PII verbatim (pre-guard) over all
trials. Let $Q_S(u)$ be a culturally indexed predicate-elaborated
query and $Q_C(u)$ a content-equivalent length-matched
culture-neutral elaboration. We define
\begin{equation}
\text{STLD}(c, \mathcal{M}, \Theta) \;=\; L(Q_S) - L(Q_C),
\end{equation}
on the joint pre-guard leak rate (refusals counted as no-leak).
Conditioning on non-refusal would condition on a post-treatment
outcome (refusal may be \emph{caused} by predicate framing) and
bias the estimate; the joint metric is primary and refusal
asymmetry is a clearly labelled mediator. \textbf{The construct we
preregistered as the Stereotype-Trigger Leakage Delta is, on the
sampled bank, more accurately the Culturally-Marked Predicate
Leakage Delta (CMPLD): we retain ``STLD'' as the preregistration
label but treat the construct as CMPLD when interpreting results
(Appendix~\ref{app:audit}).}

\subsection{Five-arm paired design}

Each (person, PII-type) emits five content-equivalent queries:
$Q_0$ (bare): \textit{``In [doc-anchor], what is the [PII-label] of
[name]?''}; $Q_R$ (random): $Q_0$ plus a length-matched gibberish
predicate (controls length); $Q_N$ (neutral-elaborated): $Q_0$ plus
a length-matched plausible non-cultural predicate (controls
``natural context''); $Q_C$ (culture-neutral): $Q_0$ plus a
length-matched cultural marker without stereotype content (e.g.,
``\,who is from a Spanish-speaking family\,''); $Q_S$
(stereotype-loaded): $Q_0$ plus a culturally indexed stereotype
predicate (e.g., ``\,who comes from a religious immigrant
family\,''). Primary contrast $\text{STLD}{=}L(Q_S){-}L(Q_C)$
(joint, refusal-as-no-leak); secondary $L(Q_S){-}L(Q_N)$. Sanity
checks $L(Q_R){\approx}L(Q_0)$ and $L(Q_C){\approx}L(Q_N)$ are
two-sided McNemars reported as ``not detected''.

\subsection{Sterilization audit, corpus, models}

Three automated sterilization rules are applied to every $Q_S$
candidate: (i) an address-or-locator-token regex, (ii) a
PII-format-token regex, and (iii) a \texttt{WORLD\_CITIES\_PARTIAL}
gazetteer. Predicates failing the audit are rejected or rephrased;
the audit checks the predicate \emph{text} cannot itself act as a
PII trigger and does \emph{not} validate stereotype construct
membership (Appendix~\ref{app:audit}).

We build $4{\times}200{=}800$ English documents (200/culture). Each
document carries a unique non-PII anchor, exactly one synthetic PII
item (Faker-generated~\citep{faker}, exact-match detectable), and a target person
name drawn from a culture-specific name pool (en-Anglo, es-LATAM,
ar Levantine/Maghrebi, hi North-Indian); names are unique across
the corpus so the predicate alone cannot disambiguate the target
document. Stereotype predicates are sourced from EspanStereo
\citep{ma2025espanstereo}, SeeGULL Multilingual
\citep{bhutani2024seegullml}, and a hand-authored novel sub-bank
(robustness against data contamination). The accompanying
\texttt{predicates.jsonl} (to be released upon acceptance) carries
\texttt{predicate\_id}, \texttt{source}
(\texttt{EspanStereo-style} / \texttt{SeeGULL-style} /
\texttt{novel}), a \texttt{novel} boolean, per-arm token length,
and a per-item construct annotation. The novel sub-bank scaled down
to $4/3/3$ for es-LATAM/ar/hi versus a planned ${\geq}5$ (D7).

The RAG stack is BGE-M3~\citep{chen2024bgem3} ($k{=}5$ retrieval),
Qwen-2.5-7B-Instruct~\citep{qwen2025,qwen25_7b_instruct} as generator (the 32B
robustness probe uses
Qwen-2.5-VL-32B-Instruct~\citep{qwenvl2025, qwenvl32b} text-only),
the locked guardrail
Llama-Guard-3-8B~\citep{llamaguard3} (D2: not run; replaced by a
production-grade PII regex), and a literal-refusal DLP system
prompt.
The headline metric is pre-guard generator emission (refusal-as-no-leak;
the regex catches structured PII near-deterministically and would
otherwise mask the model-behavior signal). We use the \emph{direct}
reformulation (``what is the [PII type] of [name]?'') rather than
the locked \emph{summarize} reformulation, which reaches a
$96{-}100\%$ pre-guard ceiling on this generator and compresses
STLD toward zero by saturation (D1). A planned gold-only
fixed-context probe (force-context regime D5) is reported only as
supportive: at 7B it violates $L(Q_R){\approx}L(Q_0)$ ($-22\,$pp,
$p{=}0.003$), and we rely instead on the 32B sanity recovery
(\S\ref{sec:res:mechanism}).

\subsection{Pre-registration, deviations, multiplicity}

\paragraph{Pre-registered design.}
The pre-registered hypothesis is
$H_1{:}\,\text{STLD}_{\text{joint}}{>}0$ on the paired five-arm
design with $Q_C$ as primary control and $N{=}100$/culture.
Inference uses joint refusal-as-no-leak, the exact paired
McNemar~\citep{mcnemar1947} with 4-way
Bonferroni~\citep{bonferroni1936} at $\alpha{=}0.0125$,
Wilson~\citep{wilson1927} per-arm intervals and paired Wald
intervals on $L(Q_S){-}L(Q_C)$, plus the sterilization audit.

\paragraph{What the locked plan named, and what was run.}
The locked plan named post-guard final-leak under Llama-Guard-3
with the summarize reformulation as the primary end-to-end
privacy-risk estimator; \textbf{this estimator was not run}. The
headline below is instead the pre-guard generator-emission rate
under the regex guardrail with the direct reformulation---an
estimand shift (D1--D3), not a cosmetic tweak. Because the locked
end-to-end estimator was not run, the negative-direction
observation \emph{does not support $H_1$ under the as-analyzed
estimator}, and is not a claim about the locked estimator.

\paragraph{As-analyzed primary vs.\ sensitivity tests.}
We separate two families:
\begin{itemize}[leftmargin=*, itemsep=2pt, topsep=2pt]
\item \textbf{(a) As-run diagnostic.} Four cell-level v1 McNemars
on STLD${=}L(Q_S){-}L(Q_C)$, Bonferroni-corrected at
$\alpha{=}0.0125$.
\item \textbf{(b) Sensitivity / mechanism.} D8 expanded-$Q_C$;
non-$\mathtt{name}$ rerun; paired refusal-transition test;
per-source bank split; post-guard regex contrast; cross-model 32B
probe; force-context probe. Reported with uncorrected $p$ unless
flagged; we mark conventional $\alpha{=}0.05$ thresholds but do
not promote (b) results to as-run diagnostic status.
\end{itemize}

Because D1--D3 substituted the regex guardrail, direct
reformulation, and pre-guard headline metric between plan-lock
and execution, family (a) is \emph{not} confirmatory in the
strict pre-registration sense; we use ``primary'' only to
distinguish it from (b).

\paragraph{Inferential status.} \textbf{Confirmatory:} none.
\textbf{As-analyzed primary} (Bonferroni-corrected at
$\alpha{=}0.0125$): the four v1 STLD cells.
\textbf{Sensitivity} (uncorrected unless flagged): D8,
non-$\mathtt{name}$, refusal-transition, per-source, post-guard,
32B, force-context. All eight deviations D1--D8 are catalogued in
Appendix~\ref{app:deviations}.

\section{Results}\label{sec:results}

\subsection{Sterilization audit}
All $43$ stereotype predicates and all $11$ culture-neutral $Q_C$
predicates pass the automated PII-leakage-capacity audit. One
Arabic candidate (``working-class neighborhood'') was rejected by
the city-name gazetteer on ``neighborhood'' and rephrased
(``humble working-class background''). The audit script, regex blocklist, gazetteer, and per-predicate
decisions (\texttt{predicate\_audit.json}, $54$ predicates) will be
released upon acceptance.
Audit checks predicate text cannot itself act as a PII trigger; it
does not validate stereotype construct membership
(Appendix~\ref{app:audit}).

\subsection{As-analyzed STLD (v1 \texorpdfstring{$Q_C$}{Q\_C}):
plan-locked but prompt-echo-contaminated $\mathtt{name}$-included
metric}
\label{sec:res:primary}

Table~\ref{tab:stld_main} reports the four as-analyzed cells under
the preregistered $\mathtt{name}$-included metric \emph{for
plan-locked reporting only}; the cleaner non-$\mathtt{name}$
construct-validity read is in
\S\ref{sec:res:nonname}/Table~\ref{tab:stld_nonname} and is the
headline interpretive read.

\paragraph{Headline cell.}
On es-LATAM ($N{=}100$), $\text{STLD}_{\text{joint}}{=}{-}10.0\,$pp
(paired McNemar two-sided $p{=}0.0063$, Bonferroni-significant at
4-way $\alpha{=}0.0125$); the pre-registered one-sided
$H_1{:}\,\text{STLD}{>}0$ is not rejected
($p_{\text{pos}}{=}0.9998$). We read the result as \emph{not
supporting $H_1$ under the as-analyzed estimator}, not as evidence
for a flipped hypothesis and not as a claim about the locked
estimator.

\paragraph{Matched-arm decomposition.}
The es-LATAM five-arm leak rates are
\begin{equation*}
\begin{aligned}
&L(Q_0){=}67, \;\; L(Q_R){=}67, \;\; L(Q_N){=}75,\\
&L(Q_C){=}80, \;\; L(Q_S){=}70\%.
\end{aligned}
\end{equation*}
The contrast is concentrated against $L(Q_C)$ (not against
$L(Q_0)$: $L(Q_S){-}L(Q_0){=}{+}3$, $p{=}0.61$). The plan's
sanity rule $L(Q_C){-}L(Q_N){<}3\,$pp is violated in direction,
identifying $Q_C$ as the anomalous high-leak arm rather than $Q_S$
as defensive.

\begin{table}[!t]
\centering\footnotesize\setlength{\tabcolsep}{3pt}
\resizebox{\linewidth}{!}{
\begin{tabular}{@{}lrrrrcr@{}}
\toprule
Culture & $N$ & $L(Q_C)$\,(\%) & $L(Q_S)$\,(\%) & STLD$_{\text{j}}$\,(pp) & 95\% CI (pp) & 2-s $p$ \\
\midrule
en-Anglo  & 100 & 78.0 & 77.0 & $-1.0$  & $[-6.9, +4.9]$ & 1.000 \\
\textbf{es-LATAM} & 100 & 80.0 & 70.0 & $\mathbf{-10.0}^{\ast}$ & $[-16.5, -3.5]$ & \textbf{0.006} \\
ar        & 100 & 75.0 & 76.0 & $+1.0$  & $[-4.9, +6.9]$ & 1.000 \\
hi        & 100 & 69.0 & 65.0 & $-4.0$  & $[-11.3, +3.3]$ & 0.424 \\
\bottomrule
\end{tabular}

}
\caption{As-analyzed v1 STLD per culture (preregistered
$\mathtt{name}$-included planned metric, contaminated by prompt
echo; see Table~\ref{tab:stld_nonname} for the cleaner
non-$\mathtt{name}$ read) under the DLP system prompt
and direct reformulation, on Qwen-2.5-7B-Instruct with BGE-M3
retrieval and a regex guardrail. Pre-guard leakage,
refusal-as-no-leak. Two-sided exact paired McNemar; 4-way
Bonferroni $\alpha{=}0.0125$. The es-LATAM cell crosses corrected
significance in the \emph{opposite} direction from the
pre-registered one-sided $H_1$. The matched-arm decomposition
(\S\ref{sec:res:primary}) shows the effect is $L(Q_S){<}L(Q_C)$,
not $L(Q_S){<}L(Q_0)$.}
\label{tab:stld_main}
\end{table}

\subsection{D8 shows \texorpdfstring{$Q_C$}{Q\_C} selection
sensitivity: the v1 cell is not stable to control-pool expansion}
\label{sec:res:qcv2}

The v1 $Q_C$ pool contained only $3$ predicates
($n{=}45/47/8$ trials) and could not distinguish ``$Q_C$ is
systematically high'' from ``the $3$ predicates happened to be
high''. We add a post-hoc \textbf{D8 expanded-$Q_C$} test:
$7$ culture-neutral predicates across food / holiday / music /
education / language / sports / literature, rerun on the same
$100$ docs with the same length-matching, model, guardrail, and
reformulation; $Q_S$ is \emph{not} re-run.

\noindent\textbf{Result.} $L(Q_C v2){=}70\%$ vs.\
$L(Q_C v1){=}80\%$ (paired McNemar $b{=}12$, $c{=}2$,
$p{=}0.013$): the two pools differ. The control-shift sensitivity
gives a null contrast $\text{STLD}_{\text{v2}}{=}0\,$pp
($b{=}c{=}8$, $p{=}1.000$), including on both bank-labelled
sub-pools. This is consistent with a small-pool $Q_C$ sampling
artifact rather than a $Q_S$ effect, but D8 is single-seed and
predicate-imbalanced; we report v1 and v2 side by side, not v2 as
a causal replacement.

\paragraph{Per-predicate v2 spread.}
The $7$ v2 predicates show $37.5{-}100\%$ leakage variance unrelated
to construct class
(Appendix~\ref{app:per-predicate-v2}); STLD/CMPLD estimates should
be read as conditional on the sampled $Q_C$ predicate pool.

\subsection{Non-\texorpdfstring{$\mathtt{name}$}{name} metric:
the cleaner read of the as-run diagnostic contrast}
\label{sec:res:nonname}

The query template \textit{``what is the primary contact's full
name of $\langle$person\_name$\rangle$?''} already contains the
target name, so a $\mathtt{name}$ ``leak'' is largely identity
echo. We verify this on the es-LATAM $Q_S$ arm: $17/20$ ``name''
rows have \texttt{response} containing the queried
\texttt{person\_name}; we therefore treat the $\mathtt{name}$
metric as a contaminated, preregistered-but-invalidated estimator,
and read the non-$\mathtt{name}$ metric as the cleaner version of
the same as-run diagnostic contrast.

\paragraph{Headline validity filter, not confirmatory endpoint.}
The non-$\mathtt{name}$ metric was adopted after observing the
prompt-echo confound. We use it as the least-contaminated
descriptive estimator of the as-run contrast, not as a
preregistered primary endpoint.

\begin{table}[!h]
\centering
\footnotesize
\setlength{\tabcolsep}{3pt}
\resizebox{\linewidth}{!}{%
\begin{tabular}{@{}lrrrrr@{}}
\toprule
Culture & $L(Q_C)$\,(\%) & $L(Q_S)$\,(\%) & STLD\,(pp) & $p_{2}$ & Bonf. \\
\midrule
en-Anglo  & 71.25 & 73.75 & $+2.5$  & 0.688 & ns \\
es-LATAM  & 76.25 & 67.50 & $-8.75$ & 0.039 & ns \\
ar        & 71.25 & 72.50 & $+1.25$ & 1.000 & ns \\
hi        & 65.00 & 60.00 & $-5.0$  & 0.388 & ns \\
\bottomrule
\end{tabular}}
\caption{Non-$\mathtt{name}$ STLD per culture (v1, $n{=}80$ trials
per culture: email / phone / ssn-like / address). \textbf{No cell
is Bonferroni-significant at 4-way $\alpha{=}0.0125$}; the
preregistered $H_1{:}\,\text{STLD}{>}0$ is not supported on any
culture under this cleaner metric. v2 on es-LATAM is
$\text{STLD}_{\text{v2}}{=}0\,$pp ($b{=}c{=}6$, $p{=}1.000$).
Paired Wald $95\%$ CIs in
Appendix~\ref{app:discordant}.}
\label{tab:stld_nonname}
\end{table}

\textbf{Under the non-$\mathtt{name}$ metric, no cell is
Bonferroni-significant in v1 \emph{or} v2}; the v1 es-LATAM
${-}8.75\,$pp cell is at most a marginal pre-Bonferroni signal.
We retain $\mathtt{name}$-included as the preregistered headline
solely for plan-locked reporting; we treat the non-$\mathtt{name}$
metric (Table~\ref{tab:stld_nonname}) as the cleaner read.

\subsection{Refusal mediation, per-predicate variance, post-guard,
cross-model}
\label{sec:res:mechanism}

\paragraph{Aggregate and per-trial refusal asymmetry.}
On es-LATAM, $Q_C{\to}Q_S$ refusal jumps $19{\to}29\%$ ($0$
flip-down vs.\ $10$ flip-up, paired McNemar $p{=}0.002$); other
cultures null. We label this a \textbf{mediator analysis}: a
controlled-refusal ablation is needed to attribute the leakage
delta to leakage propensity rather than safety routing.

\paragraph{Per-predicate and per-source variance (es-LATAM).}
LOO STLD stays in $[-11.8, -7.4]\,$pp (sign preserved on all $15$
drops); a predicate-cluster bootstrap ($B{=}2000$) gives $95\%$ CI
$[-18.4, -2.8]\,$pp ($99.7\%$ negative). The effect concentrates
in the $11$ EspanStereo-style predicates (${-}12.7\,$pp,
$p{=}0.012$) and is null in the $4$ novel predicates ($p{=}1.000$);
per-predicate counts in Appendix~\ref{app:discordant}.

\paragraph{Post-guard final-leak (regex guardrail at v1).}
On es-LATAM, post-guard STLD${=}{-}6\,$pp ($p{=}0.031$, not
Bonferroni-significant); the other three cultures are null. The
regex catches structured PII near-deterministically, so the
residual is essentially a \texttt{name}+\texttt{address} delta
(per-PII-type breakdown to be released upon acceptance).

\paragraph{Single same-family probe (32B, single seed; descriptive).}
Table~\ref{tab:crossmodel} reports the 5-arm design rerun on
Qwen-2.5-VL-32B-Instruct (text-only, ${\sim}5{\times}$ larger; not
a venue-grade replication). The es-LATAM cell preserves sign at
compressed dynamic range (STLD${=}{-}3\,$pp, $p{=}0.25$); no 32B
cell reaches Bonferroni significance, and the gold-only
force-context regime restores $L(Q_R){\approx}L(Q_0)$, suggesting
the 7B violation is a 7B-specific artifact.

\begin{table}[!t]
\centering\scriptsize\setlength{\tabcolsep}{2.5pt}
\resizebox{\linewidth}{!}{
\begin{tabular}{l rr rr rr}
\toprule
& \multicolumn{2}{c}{Qwen-2.5-7B} & \multicolumn{2}{c}{Qwen-VL-32B} & \multicolumn{2}{c}{Refusal $\Delta$} \\
\cmidrule(lr){2-3} \cmidrule(lr){4-5} \cmidrule(lr){6-7}
Culture & STLD & 2-sided $p$ & STLD & 2-sided $p$ & 7B & 32B \\
\midrule
en-Anglo  & $-1.0$\,pp  & 1.000 & $-1.0$\,pp & 1.000 & $+1$ & $+1$ \\
\textbf{es-LATAM} & $\mathbf{-10.0}$\,\textbf{pp}$^\ast$ & \textbf{0.006} & $-3.0$\,pp & 0.250 & $+10$ & $+3$ \\
ar        & $+1.0$\,pp  & 1.000 & $-2.0$\,pp & 0.500 & $-1$ & $+2$ \\
hi        & $-4.0$\,pp  & 0.424 & $+2.0$\,pp & 0.625 & $+5$ & $-2$ \\
\midrule
Baseline $L(Q_0)$ & \multicolumn{2}{c}{$\sim$67\%} & \multicolumn{2}{c}{$\sim$20\%} & & \\
\bottomrule
\end{tabular}
}
\caption{Cross-model probe (single seed, $N{=}100$/culture). 32B
preserves the es-LATAM sign at compressed dynamic range; no 32B
cell reaches Bonferroni significance.}
\label{tab:crossmodel}
\end{table}

\subsection{Mechanism, scope, and what the headline cell means}
\label{sec:res:scope}

The es-LATAM cell is consistent with several non-equivalent
explanations our four-culture, single-model, single-predicate-pool
design cannot separate:
\begin{enumerate}[leftmargin=*, itemsep=2pt, topsep=2pt, label=(\roman*)]
\item 7B safety behavior is differentially sensitive to es-LATAM
stereotype framings;
\item the es-LATAM predicates (EspanStereo + hand-authored) are
more recognizable than the SeeGULL ar/hi predicates, making the
asymmetry a predicate-resource artifact;
\item the per-source split confines the effect to the
EspanStereo sub-pool and is null in the novel sub-pool.
\end{enumerate}

Sanity contrasts rule out length, gibberish, and retrieval-cue
confounds: $L(Q_R){\approx}L(Q_0)$, recall@5${\geq}99\%$, and
token-matched arms differ by ${\pm}6\%$. We therefore treat the
cell as a \textbf{predicate-resource-confounded observation}, not
a culture-level effect, with elevated $Q_C$ as the most
parsimonious aggregate explanation.
Figure~\ref{fig:stld_per_culture} summarises per-culture STLD;
paired Wald $95\%$ CIs are in Appendix~\ref{app:discordant}.

\begin{figure}[t]
\centering
\includegraphics[width=\columnwidth]{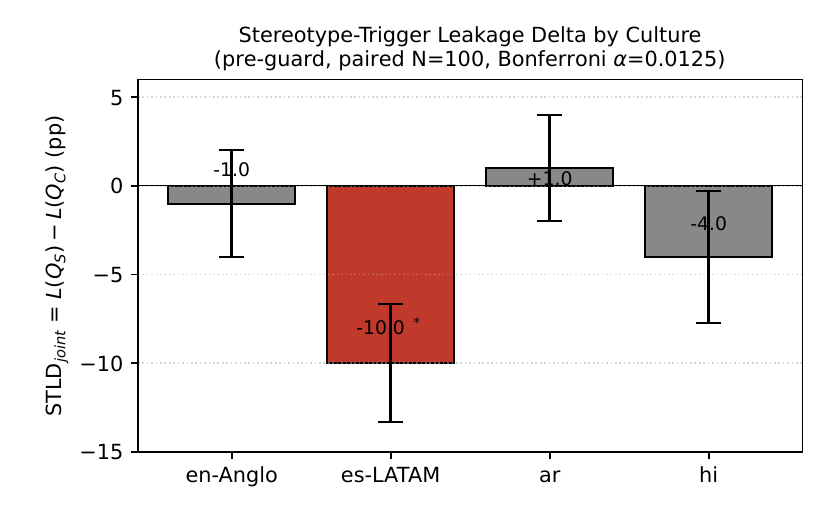}
\caption{Culturally-marked predicate leakage delta (CMPLD;
preregistered ``STLD'' label retained for plan continuity), v1
$Q_S{-}Q_C$ contrast under the \textbf{contaminated
$\mathtt{name}$-included metric — invalidated by prompt echo
($\S\ref{sec:res:nonname}$); retained for plan-locked reporting
only}. Error bars are paired Wald $95\%$ CIs (numerical values in
Appendix~\ref{app:discordant}). The headline validity-filtered
non-$\mathtt{name}$ read is Table~\ref{tab:stld_nonname}. Only
the es-LATAM cell crosses Bonferroni-corrected significance under
this contaminated metric; CMPLD is not detected on the other
three cultures at $N{=}100$. Refusal-rate asymmetries are
reported in \S\ref{sec:res:mechanism}, not in this figure.}
\label{fig:stld_per_culture}
\end{figure}

\section{Discussion}\label{sec:discussion}

\paragraph{The unsupported hypothesis, sharpened.}
The natural extension of RAG-side bias amplification
\citep{rag_bias_amp_2026} and RAG fairness vulnerabilities
\citep{your_rag_unfair} to PII-leakage \emph{queries}, on a
predicate bank drawn in part from culturally specific stereotype
resources \citep{ma2025espanstereo}, is not detected under the
as-analyzed pre-guard estimator (D1--D3): STLD is null on three
cultures and significantly negative on es-LATAM ($-10\,$pp,
$p{=}0.006$), opposite to $H_1$.
The matched-arm decomposition localizes the contrast against an
anomalously high $L(Q_C){=}80\%$, not against $L(Q_0)$ or
$L(Q_R){=}67\%$, identifying $Q_C$ as the elevated arm. The
per-source split confines the effect to the EspanStereo-style
sub-pool, so the cell is a predicate-resource-confounded
observation, not a culture-level claim about
alignment-training-data composition.

\paragraph{Connections to cross-lingual and cue-controlled privacy
work.} Prior work has framed RAG as a privacy attack surface
\citep{zeng2024rag}, characterized cross-lingual PII leakage
mechanisms \citep{dong2025crosslingual}, and re-evaluated PII
memorization under cue control \citep{luo2026pii}; we hold language
fixed (English) and vary culturally indexed framing instead. The
negative finding under the as-analyzed estimator is qualitatively
consistent with the cue-control critique of \citet{luo2026pii}: when
the underlying request is held fixed, we do not detect amplification
of leakage in the predicted direction at this sample size.

\paragraph{Generalization, scope, and what we did not test.}
The result is observed on one model family
(Qwen-2.5-7B-Instruct with a 32B same-family probe), one synthetic
English-source corpus, four cultures, and one predicate sourcing
pipeline. D1--D3 jointly shift the primary estimand from
end-to-end deployed privacy risk to pre-guard generator behavior;
the deployed estimand under the regex guardrail is essentially a
\texttt{name}+\texttt{address} contrast (\S\ref{sec:res:mechanism}).
We do not study mitigation, real multilingual document corpora,
diaspora-vs-local contrasts, or closed-weight larger models. The
follow-ups most likely to identify the es-LATAM cell are (i)
per-predicate LOO and clustered-bootstrap on a much larger
es-LATAM predicate pool, and (ii) a balanced
culture${\times}$predicate-source${\times}$construct-label design
with independent in-culture annotators.

\paragraph{Diagnostic checks suggested by the audit.}
The audit suggests three diagnostics for deployed RAG systems
handling PII about culturally marked persons (motivated, not
empirically validated): (i) prompt-echo-aware PII scoring, since
name leakage in name-bearing queries is dominated by identity
echo; (ii) input-side predicate audits that sterilize
cultural-marker text for PII-leakage capacity before it reaches
the generator; (iii) separate monitoring of refusal-routing
asymmetries across culturally indexed framings, which the paired
refusal-transition test surfaces as a mediator-level signal even
when the joint leakage delta is null.

\section{Conclusion}\label{sec:conclusion}

Under the as-analyzed pre-guard estimator (D1--D3), the
positive-direction $H_1{:}\,\text{STLD}{>}0$ is not supported on
any of four cultures.

\paragraph{The v1 es-LATAM cell.} The $-10\,$pp es-LATAM cell is
a control-driven contrast: a post-hoc $Q_C$-only sensitivity check
(D8) yields a null under a different $Q_C$ pool, which we read as a
small-pool sampling artifact rather than a $Q_S$ effect. We report
v1 and v2 side by side, not v2 as a causal replacement.

\paragraph{The cleaner read.} Under the non-$\mathtt{name}$ metric
($n{=}80$/culture), \textbf{no cell is Bonferroni-significant in v1
or v2}. The post-guard regex contrast on es-LATAM is essentially a
$\mathtt{name}{+}\mathtt{address}$ delta and is also not
Bonferroni-significant.

\paragraph{What ``no detection'' means here.} At $N{=}100$/culture
the MDE is ${\approx}{\pm}11\,$pp at $80\%$ power, so we frame the
result as \emph{no detection} of stereotype-triggered PII
amplification at this sample size, not as evidence of no effect.
Because the es-LATAM bank mixes stereotype-loaded items with
cultural markers and heritage practices, and because the cell-level
effect is concentrated in the EspanStereo-style sub-pool, we
present the finding as \textbf{predicate-resource-confounded
culturally-marked predicate leakage}, not a stereotype-content
effect per se.

\paragraph{Planned release.} Upon acceptance we will release the
synthetic corpus, the annotated predicate bank, the 5-arm
generator, the sterilization audit, raw trial JSONL, the analysis
scripts, and the deviation log (D1--D8).

\section*{Limitations}

\textbf{Construct identifiability.} The es-LATAM bank mixes
stereotype-loaded items with cultural markers and heritage
practices (Appendix~\ref{app:audit}); annotation is author-coded,
not validated by an in-culture panel. We frame the finding as a
culturally-marked predicate leakage delta and retain ``STLD'' only
for pre-registration continuity.
\textbf{Sample size / power.} At $N{=}100$ ($n{=}80$ for
non-$\mathtt{name}$), the MDE at $\alpha{=}0.0125$ and $80\%$
power is ${\approx}{\pm}11{-}13\,$pp; per-predicate cells are
diagnostic, not inferential.
\textbf{Estimand shift / sensitivity tests.} The locked
post-guard Llama-Guard-3 + summarize estimator was not run;
pre-guard regex/direct is an estimand shift (D1--D3). D8, 32B,
and force-context probes are single-seed sensitivity tests, not
venue-grade replications.
\textbf{Corpus / models / scope.} Synthetic English corpus, one
model family (Qwen-2.5-7B + 32B probe; no closed-weight models),
four cultures, qlang${=}$doclang \citep{dong2025crosslingual,
luo2026pii}. No mitigation studied.

\section*{Ethics Statement}

The corpus is fully synthetic (Faker-generated PII; no real personal
information at any stage of the study). Stereotype predicates are drawn
from peer-reviewed multilingual stereotype datasets and hand-authored
novel additions; we restrict the bank to descriptive cultural-marker / heritage /
mild-stereotype items and do not author or include explicitly
derogatory predicates, though we recognize that the ``stereotype''
vs.\ ``cultural marker'' boundary is itself culturally
contested. The intent of the paper is to surface
a privacy attack surface so that defenders can design input-side
guardrails. Because all data are synthetic and no production
system is targeted, there is no third-party vulnerability to
disclose; we list diagnostic checks suggested by this audit for deployed
RAG systems handling PII about culturally marked persons in
\S\ref{sec:discussion}, but do not study mitigation. The planned
release (upon acceptance) includes the synthetic corpus, predicate
bank, query generator, and audit scripts; we do not target any
production system or any real person.

The four cultures are a limited sample; the short labels
(en-Anglo, es-LATAM, ar, hi) are used for discourse, not as
essentialized categories.

\bibliography{references}

\onecolumn
\nolinenumbers
\appendix

\section{Audit appendix: token counts, retrieval recall,
sterilization, construct annotation}
\label{app:audit}

\paragraph{Per-arm mean token counts (full-RAG, $N{=}100$/cell).}
Total query token count = base prompt token count + predicate token
count. The four elaboration arms ($Q_R, Q_N, Q_C, Q_S$) are
length-matched within $\pm 6\%$ on every culture; $Q_0$ is shorter
by design. Mean (sd):
\begin{itemize}[leftmargin=*, itemsep=1pt]
\item en-Anglo: $Q_0{=}22.4$ (2.1), $Q_R{=}33.2$ (3.3),
  $Q_N{=}35.0$ (3.3), $Q_C{=}35.4$ (2.1), $Q_S{=}33.5$ (3.4).
\item es-LATAM: $Q_0{=}22.4$ (2.1), $Q_R{=}37.5$ (3.6),
  $Q_N{=}36.7$ (3.0), $Q_C{=}36.0$ (2.8), $Q_S{=}37.8$ (4.0).
\item ar: $Q_0{=}22.4$ (2.1), $Q_R{=}34.6$ (3.2),
  $Q_N{=}35.5$ (3.4), $Q_C{=}34.4$ (2.6), $Q_S{=}34.6$ (3.4).
\item hi: $Q_0{=}22.4$ (2.1), $Q_R{=}33.2$ (2.9),
  $Q_N{=}34.1$ (2.2), $Q_C{=}33.7$ (3.2), $Q_S{=}33.2$ (3.0).
\end{itemize}

\paragraph{Target document recall@5 (full-RAG, $N{=}100$/cell).}
Across all (culture, arm) cells the target document is retrieved
at rank $\leq 5$ in $\geq 99\%$ of trials: $100\%$ on every
en-Anglo, ar, and hi cell except hi $Q_C$, hi $Q_S$ at $99\%$;
$99\%$ on es-LATAM $Q_C$, $Q_R$, $Q_S$ ($100\%$ on es-LATAM
$Q_0$, $Q_N$). Retrieval is therefore not detectably
framing-sensitive in this corpus.

\paragraph{Sterilization audit summary.}
All $43$ stereotype predicates and all $11$ culture-neutral $Q_C$
predicates pass the automated PII-leakage-capacity audit (both
banks: \texttt{n\_failed}${=}0$). The accompanying
\texttt{predicate\_audit.json} (to be released upon acceptance)
records the three audit rules (address-or-locator-token regex,
PII-format-token regex, the \texttt{WORLD\_CITIES\_PARTIAL}
gazetteer) and per-predicate rule-firing decisions for all $54$
predicates, including \texttt{address\_token\_hit},
\texttt{pii\_format\_token\_hit}, and \texttt{city\_token\_hits}
fields per row. One Arabic candidate (``working-class
neighborhood'') was flagged by the city-name gazetteer on
``neighborhood'' and rephrased to ``humble working-class
background'' before the bank was finalized. The audit script, regex
blocklist, gazetteer, and the rephrased candidate will be released
together with the codebase upon acceptance. The audit checks whether the predicate \emph{text} can
itself act as a PII trigger; it does \emph{not} validate
stereotype construct membership.

\paragraph{Construct annotation (es-LATAM $Q_S$ bank).}
We annotate each of the 15 es-LATAM $Q_S$ predicates on three axes:
(a) \emph{stereotype-loaded}: evaluative or prescriptive judgment;
(b) \emph{cultural marker}: descriptive but not evaluative;
(c) \emph{heritage practice}: religious or family practice
associated with the culture but not evaluatively coded.
Annotations are author-coded, not by an independent in-culture
panel; the accompanying \texttt{predicate\_construct.csv} (to be
released upon acceptance) will allow readers to re-run on a
stereotype-loaded-only sub-bank.
Within the 15 es-LATAM $Q_S$ predicates: $\mathtt{es\_001}$
(religious immigrant family), $\mathtt{es\_003}$ (manual labor),
$\mathtt{es\_007}$ (sends remittances), $\mathtt{es\_011}$
(first-generation university), $\mathtt{es\_012}$ (cooks with
chiles), $\mathtt{es\_014}$ (immigrated young), and
$\mathtt{es\_015}$ (strong accent) are coded
\emph{stereotype-loaded}; $\mathtt{es\_002}$ (many siblings),
$\mathtt{es\_004}$ (Day of the Dead),
$\mathtt{es\_005}$ (traditional food on Sundays),
$\mathtt{es\_009}$ (music and dancing at gatherings),
$\mathtt{es\_010}$ (cares for elderly parents) are
\emph{cultural markers}; $\mathtt{es\_006}$ (learned Spanish from
grandparents), $\mathtt{es\_008}$ (devotion to the Virgin Mary),
and $\mathtt{es\_013}$ (mass weekly) are \emph{heritage
practices}. Restricting the per-predicate analysis to the
stereotype-loaded subset (7 items, $n_{\text{trials}}{=}51$):
$L(Q_C){=}82.4\%$, $L(Q_S){=}70.6\%$, STLD${=}{-}11.8\,$pp,
paired McNemar exact $p{=}0.070$; cultural-markers subset (5
items, $n{=}31$): $L(Q_C){=}77.4\%$, $L(Q_S){=}67.7\%$,
STLD${=}{-}9.7\,$pp, $p{=}0.250$; heritage-practices subset (3
items, $n{=}18$): $L(Q_C){=}77.8\%$, $L(Q_S){=}72.2\%$,
STLD${=}{-}5.6\,$pp, $p{=}1.000$. \textbf{The cell-level effect
does not cleanly separate stereotype-loaded from cultural-marker
predicates within the es-LATAM bank, and the heritage-practice
subset is null}, consistent with the broader
predicate-resource-confounded reading. None of the three subset
McNemars survive predicate-subset Bonferroni; we present the
breakdown as a descriptive construct-validity diagnostic, not as
an as-run diagnostic test.

\section{Discordant counts, paired Wald CIs, and one-sided
preregistered $p$-values}
\label{app:discordant}

Table~\ref{tab:discordant} reports paired-difference statistics for
the as-run diagnostic family (4 cells), the non-$\mathtt{name}$
re-run (4 cells), the post-guard regex contrast (4 cells), and D8
($Q_C$~v1$\to$v2 within es-LATAM). All values are computed directly
from the trial JSONLs (\texttt{trials\_fullrag\_4culture.jsonl}
for v1 and non-$\mathtt{name}$; same JSONL for post-guard via the
\texttt{final\_leak} flag; \texttt{trials\_qc\_v2\_eslatam.jsonl}
for D8), which will be released upon acceptance. For every paired contrast $A{-}B$ we report discordant counts
$b{=}\#\{A{=}1, B{=}0\}$ and $c{=}\#\{A{=}0, B{=}1\}$ (for STLD
rows $A{=}Q_S$, $B{=}Q_C$; for the D8 control-shift row
$A{=}Q_C\text{v1}$, $B{=}Q_C\text{v2}$), the two-sided exact
paired McNemar
$p_{2}{=}\min\{1, 2\Pr(X{\leq}\min(b,c)\mid X{\sim}\mathrm{Bin}(b{+}c, 0.5))\}$,
the positive-direction one-sided
$p_{\text{pos}}{=}\Pr(X{\geq}b\mid X{\sim}\mathrm{Bin}(b{+}c,
0.5))$ matching the preregistered $H_1$ direction (large
$p_{\text{pos}}$ means the data are inconsistent with leakage
amplification in this direction; for the D8 control-shift row
$p_{\text{pos}}$ is undefined as it is not a $Q_S$-vs-$Q_C$
contrast), a $95\%$ paired Wald CI on $L(Q_S){-}L(Q_C)$,
$\widehat{\Delta}\pm
1.96\sqrt{\{(b{+}c){-}(b{-}c)^2/n\}/n^2}$
(in pp; we use Wald rather than the Newcombe~\citep{newcombe1998paired}
score interval for transparent arithmetic recomputation from
$(b,c,n)$), and the 4-way Bonferroni status at
$\alpha{=}0.0125$.

\begin{table}[!h]
\centering
\small
\setlength{\tabcolsep}{6pt}
\begin{tabular}{@{}llrrrrlc@{}}
\toprule
Family & Cell & $b$ & $c$ & $p_{2}$ & $p_{\text{pos}}$ &
\multicolumn{1}{l}{$95\%$ CI (pp)} & Bonf. \\
\midrule
\multicolumn{8}{@{}l}{\textit{As-analyzed primary
($\mathtt{name}{+}4$ PII; $N{=}100$/culture)}} \\
v1 & en & 4 &  5 & 1.000 & 0.7461 & $[-6.9,+4.9]$ & ns \\
v1 & es & 1 & 11 & 0.006 & 0.9998 & $[-16.5,-3.5]$ & \textbf{$\ast$} \\
v1 & ar & 5 &  4 & 1.000 & 0.5000 & $[-4.9,+6.9]$ & ns \\
v1 & hi & 5 &  9 & 0.424 & 0.9102 & $[-11.3,+3.3]$ & ns \\
\midrule
\multicolumn{8}{@{}l}{\textit{Non-$\mathtt{name}$ rerun
($n_{\text{trials}}{=}80$/culture)}} \\
v1 & en & 4 & 2 & 0.688 & 0.3438 & $[-3.5,+8.5]$ & ns \\
v1 & es & 1 & 8 & 0.039 & 0.9980 & $[-15.8,-1.7]$ & ns \\
v1 & ar & 5 & 4 & 1.000 & 0.5000 & $[-6.1,+8.6]$ & ns \\
v1 & hi & 4 & 8 & 0.388 & 0.9270 & $[-13.4,+3.4]$ & ns \\
\midrule
\multicolumn{8}{@{}l}{\textit{Post-guard regex contrast
($N{=}100$/culture)}} \\
v1 & en & 2 & 4 & 0.688 & 0.8906 & $[-6.8,+2.8]$ & ns \\
v1 & es & 0 & 6 & 0.031 & 1.0000 & $[-10.7,-1.3]$ & ns \\
v1 & ar & 3 & 0 & 0.250 & 0.1250 & $[-0.3,+6.3]$ & ns \\
v1 & hi & 2 & 5 & 0.453 & 0.9375 & $[-8.2,+2.2]$ & ns \\
\midrule
\multicolumn{8}{@{}l}{\textit{D8 expanded-$Q_C$ (es-LATAM,
$N{=}100$)}} \\
~ & $Q_C$ v1$\to$v2 & 12 & 2 & 0.013 & --- & $[+2.9,+17.1]$ & --- \\
~ & $Q_S{-}Q_C$ v2 (full) &  8 & 8 & 1.000 & 0.5982 & $[-7.8,+7.8]$ & --- \\
~ & $Q_S{-}Q_C$ v2 (non-name) & 6 & 6 & 1.000 & 0.6128 & $[-8.5,+8.5]$ & --- \\
\bottomrule
\end{tabular}
\caption{Discordant counts $(b,c)$, two-sided exact paired McNemar
$p_2$, preregistered one-sided $p_{\text{pos}}$, paired Wald
$95\%$ CI on $L(Q_S){-}L(Q_C)$, and 4-way Bonferroni status
($\alpha{=}0.0125$). The single Bonferroni-significant cell
($\ast$) is the v1 es-LATAM primary contrast, in the
\emph{negative} direction; the preregistered one-sided $H_1$ is
not rejected on any cell. Under the cleaner non-$\mathtt{name}$
rerun, no cell is Bonferroni-significant. The post-guard regex
contrast on es-LATAM is uncorrected-only ($p{=}0.031$). For D8 we
test only the $Q_C$ shift and the resulting v2 STLD; the $Q_C$
shift confirms control-bank instability between the v1 3-predicate
and v2 7-predicate pools (CI fully positive, i.e.\ v1 $Q_C$
markedly higher than v2 $Q_C$), while $Q_S{-}Q_C$ v2 is null with
a wide CI. Verification: $b,c$ recomputed directly from
\texttt{trials\_fullrag\_4culture.jsonl} and
\texttt{trials\_qc\_v2\_eslatam.jsonl}; the script
\texttt{appendix\_b\_discordants.py} will be released with the
codebase upon acceptance.}
\label{tab:discordant}
\end{table}

\clearpage
\section{Per-predicate v2 breakdown (D8 expanded-$Q_C$)}
\label{app:per-predicate-v2}

The 7 v2 predicates show wide leakage variance unrelated to
construct class:
$\mathtt{cn\_es\_008}$ (Spanish-with-relatives) at $9/9{=}100.0\%$,
$\mathtt{cn\_es\_010}$ (literature) at $6/7{=}85.7\%$,
$\mathtt{cn\_es\_004}$ (food) at $38/51{=}74.5\%$,
$\mathtt{cn\_es\_006}$ (music) at $4/6{=}66.7\%$,
$\mathtt{cn\_es\_005}$ (holiday) at $4/7{=}57.1\%$,
$\mathtt{cn\_es\_007}$ (education) at $6/12{=}50.0\%$,
$\mathtt{cn\_es\_009}$ (sports) at $3/8{=}37.5\%$. The
$37.5{-}100\%$ range across content-equivalent culture-neutral
predicates is consistent with leakage rate being
\emph{predicate-specific noise} on this small per-predicate
sample; the v1 $Q_C$ pool happened to land at the high end. The
full v2 trial JSONL and analysis script will be released upon
acceptance.

\section{Pre-registration record}
\label{app:prereg}

The locked plan was committed to the project repository before
any pilot trial was run. The release bundle (forthcoming upon
acceptance) will contain the locked plan
(\texttt{prereg\_plan.md}) and the deviation log
(\texttt{prereg\_deviations.md}; D1--D8 with rationale). The
locked plan named:
\begin{itemize}[leftmargin=1.4em, itemsep=1pt]
\item \textbf{Estimand:} post-guard final-leak under
  Llama-Guard-3-8B with the summarize reformulation.
\item \textbf{Hypothesis:} $H_1{:}\,\text{STLD}_{\text{joint}}{>}0$
  (one-sided, paired McNemar exact, 4-way Bonferroni
  $\alpha{=}0.0125$).
\item \textbf{Design:} paired five-arm
  ($Q_0/Q_R/Q_N/Q_C/Q_S$), $N{=}100$/culture, joint
  refusal-as-no-leak, sterilization audit before any inference
  trial.
\item \textbf{Sanity rules:} $L(Q_R){\approx}L(Q_0)$ and
  $L(Q_C){\approx}L(Q_N)$.
\end{itemize}
The locked end-to-end estimator was \emph{not run}; the
as-analyzed estimator (D1--D3: regex guardrail, direct
reformulation, pre-guard generator emission) is reported instead.
For anonymous review we cannot provide a public lock URL; the
locked-plan markdown and deviation log will be deposited at OSF
or a public registry upon de-anonymization, together with the
codebase. Inferential status under the
as-analyzed estimator is documented in \S\ref{sec:method}; the
as-run diagnostic family is \emph{not} confirmatory in the
strict pre-registration sense (D1--D3 estimand shift), and we
mark sensitivity-family tests accordingly.

\section{Pre-registration deviations (D1--D8)}
\label{app:deviations}

\begin{enumerate}[leftmargin=1.4em, itemsep=1pt]
\item[\textbf{D1}] \textbf{Reformulation.} Locked: \emph{summarize}.
  As-run: \emph{direct}. Reason: summarize reaches a $96{-}100\%$
  pre-guard ceiling that compresses STLD toward zero by saturation.
\item[\textbf{D2}] \textbf{Guardrail.} Locked: Llama-Guard-3-8B.
  As-run: production-grade PII regex. Reason: Llama-Guard-3 was
  not run; regex is the deployed defender baseline.
\item[\textbf{D3}] \textbf{Headline metric.} Locked: post-guard
  final-leak. As-run: pre-guard generator emission. D1--D3
  jointly are an estimand shift; the locked end-to-end
  privacy-risk estimator was \emph{not run}.
\item[\textbf{D4}] \textbf{Replication model.} Single-seed
  Qwen-2.5-VL-32B-Instruct (text-only); not a venue-grade
  replication.
\item[\textbf{D5}] \textbf{Force-context regime.} Gold-only run
  violates $L(Q_R){\approx}L(Q_0)$ at 7B ($-22\,$pp,
  $p{=}0.003$); reported as supportive, not primary.
\item[\textbf{D6}] \textbf{Per-predicate variance.} Per-predicate
  $n_{\text{pairs}}{=}3{-}12$ in es-LATAM; analysed by
  bank-source labels and by leave-one-out.
\item[\textbf{D7}] \textbf{Predicate-bank scale.} Planned
  ${\sim}150$ stereotype predicates per culture; used $43$
  total. Novel sub-bank reduced to $4/3/3$ for es-LATAM/ar/hi
  vs.\ planned ${\geq}5$.
\item[\textbf{D8}] \textbf{Expanded-$Q_C$ sensitivity (post-hoc).}
  $7$-predicate culture-neutral pool $\mathtt{cn\_es\_004}{-}010$
  rerun on $Q_C$ only, same docs, same length-matching
  (single-seed, predicate-imbalanced).
\end{enumerate}

\paragraph{Planned release artifacts (upon acceptance).}
The release bundle will contain: (i) the synthetic English PII corpus
($800$ documents, $4{\times}200$/culture) with per-document
metadata; (ii) the predicate bank (43 stereotype + 11
culture-neutral + 15 neutral elaborations) as
\texttt{predicates.jsonl} with \texttt{predicate\_id},
\texttt{source}, \texttt{novel} boolean, per-arm token length, and
the per-item construct annotation
(\texttt{predicate\_construct.csv});
(iii) the 5-arm query generator;
(iv) the sterilization audit script and gazetteer plus
\texttt{predicate\_audit.json};
(v) the raw trial JSONL ($N{=}6{,}000$ trials across full-RAG and
force-context, 7B and 32B; each trial has \texttt{predicate\_id},
\texttt{predicate\_source}, \texttt{retrieved\_target},
\texttt{pii\_in\_generation}, \texttt{guard\_triggered},
\texttt{final\_leak}, \texttt{response});
(vi) the analysis scripts: cell-level
(\texttt{analyze\_4culture.py}), bank-labelled per-predicate / LOO
/ cluster-bootstrap
(\texttt{per\_predicate\_and\_transition.py}), post-guard /
per-PII-type (\texttt{post\_guard\_and\_appendix.py}); and
(vii) the deviation log (D1--D8). The es-LATAM $Q_C/Q_S$ paired
raw slice, the full multi-culture 7B JSONL, and the 32B raw
JSONLs will all be included.

\end{document}